\documentclass{article}

\usepackage[final]{neurips_2023}

\usepackage[utf8]{inputenc} % allow utf-8 input
\usepackage[T1]{fontenc}    % use 8-bit T1 fonts
\usepackage{hyperref}       % hyperlinks
\usepackage{url}            % simple URL typesetting
\usepackage{booktabs}       % professional-quality tables
\usepackage{amsfonts}       % blackboard math symbols
\usepackage{nicefrac}       % compact symbols for 1/2, etc.
\usepackage{microtype}      % microtypography
\usepackage{xcolor}         % colors
\usepackage{quoting,xparse}
\usepackage{multirow}
\usepackage{graphicx}
\usepackage{multirow}
\usepackage{subcaption}
\usepackage{amsmath}

\NewDocumentCommand{\bywhom}{m}{% the Bourbaki trick
  {\nobreak\hfill\penalty50\hskip1em\null\nobreak
   \hfill\mbox{\normalfont(#1)}%
   \parfillskip=0pt \finalhyphendemerits=0 \par}%
}

\NewDocumentEnvironment{pquotation}{m}
{\begin{quoting}[
    indentfirst=true,
    leftmargin=97pt, % 97pt 대신 em 단위를 사용하는 것이 더 유연합니다.
    rightmargin=\parindent]\itshape}
{\bywhom{#1}\end{quoting}}

\title{Deeper Inside Deep ViT}

\author{%
  Sungrae Hong\\
  Graduate School of Data Science\\
  KAIST\\
  DaeJeon, Republic of Korea\\
  \texttt{sun.hong@kaist.ac.kr} \\
}

\begin{document}

\maketitle

\begin{abstract}
There have been attempts to create large-scale structures in vision models similar to LLM, such as ViT-22B. While this research has provided numerous analyses and insights, our understanding of its practical utility remains incomplete. Therefore, we examine how this model structure reacts and train in a local environment. We also highlight the instability in training and make some model modifications to stabilize it. The ViT-22B model, trained from scratch, overall outperformed ViT in terms of performance under the same parameter size. Additionally, we venture into the task of image generation, which has not been attempted in ViT-22B. We propose an image generation architecture using ViT and investigate which between ViT and ViT-22B is a more suitable structure for image generation.\footnote{Source code is available at \url{https://github.com/HongSungRae/ai602}} \\
\end{abstract}

\section{Introdcution}

\begin{pquotation}{Hyejeong Choi, Netflix series \texttt{The Glory}, 2022}
"No, I told you I like big ones. I like everything big."\\
\end{pquotation}
% \begin{pquotation}{Hyejeong Choi, Netflix series \texttt{The Glory}, 2022}
% "No, I told you I like big ones. I like everything big."
% \end{pquotation}
% \begin{quote}
%   "No, I told you I like big ones. I like everything big." \\
%   Hyejeong Choi, Netflix series \texttt{The Glory}, 2022
% \end{quote}

The advent of the Largest Large Language Models (LLMs)~\cite{chowdhery2023palm,thoppilan2022lamda} has provided a new perspective on how we deal with deep learning models. It has demonstrated superior performance by designing a massive foundation model first and then applying it to downstream tasks, as opposed to creating smaller task-specific models~\cite{moor2023foundation,wang2023internimage}. This trend is increasingly prominent in vision research as well. \cite{zhai2022scaling} proposed a method for scaling the ViT model, showcasing the feasibility of envisioning a foundation model for vision. However, this attempt, with a size of 15B, is still small compared to LLM models with 540B~\cite{chowdhery2023palm}. Therefore, ViT-22B~\cite{dehghani2023scaling} elevates the number of learnable parameters to 22B through architectural and recipe modifications. This model not only achieved state-of-the-art or comparable performance in numerous quantitatively evaluable downstream tasks but also demonstrated superior performance qualitatively compared to several vision models.

In this manner, we are seeking findings aligned with the belief that larger sizes of all deep learning models would be beneficial. However, some studies caution against the belief that a foundation model is superior for all tasks, emphasizing the need for careful consideration~\cite{bommasani2021opportunities}. This has been analyzed from a linguistic perspective, where comments on the model's reliability and potential for development vary among experts in different fields. Due to the slower pace of progress in vision foundation models compared to language foundation models, it is challenging to pinpoint specific issues. However, what is certain is that ViT-22B is trained on a massive private dataset~\cite{sun2017revisiting,mehta2022large}, and its trained parameters are not disclosed\footnote{\url{https://www.reddit.com/r/MachineLearning/comments/9zj7wl/d_how_do_i_get_googles_jft300m_dataset/}}, making it difficult for us to derive benefits from it. This implies that if we have data available at the local or station level, we need to train the ViT-22B model ourselves. Moreover, we are yet to understand whether the model's advancements solely originate from its size or if there are reasons embedded in its structure changes. Existing research lacks experiments that unify the size of the model and the pre-training dataset(\textit{i.e. }JFT). Therefore, we locally train the ViT-22B structure and observe the training process to understand how it reacts. During the model's training, we observed a phenomenon of the gradient explosively increasing. This appears to stem from the parallel linear structure without normalization or regulation. As evidence, the introduction of normalization to this parallel network resolved the issue. The results of scratch training ViT and ViT-22B under the same conditions guide us in choosing the appropriate structure.

Additionally, we analyze the task of image generation, which has not been attempted in the ViT-22B research. While image generation is a widely studied area in computer vision~\cite{liu2021blendgan, ramesh2021zero,chen2021generative,yi2019generative}, there is limited research based on the ViT architecture~\cite{lee2021vitgan,jiang2021transgan,esser2021taming}. We introduce a new structure called ViTUnet and use ViT and ViT-22B as backbones to observe how these models perform in image generation. The ViTUnet structure involves a process similar to Unet~\cite{ronneberger2015u}, incorporating residual and encoding-decoding steps to map and reconstruct images in lower dimensions. We highlight the absence of an image reconstruction structure based on ViT and present the performance of our proposed model qualitatively and quantitatively.

% On the Opportunities and Risks of Foundation Models
% 추가적으로 우리는 ViT-22B 모델에서 시도되지 않았던 task를
\section{Related Work}

\subsection{Vision Transformer}

The Transformer architecture~\cite{vaswani2017attention} has achieved remarkable performance improvements in language translation~\cite{fuchs2020se} and question-answering tasks~\cite{chowdhery2023palm}. The operations based on attention mechanisms faithfully mimic language capabilities that require deducing relationships between words in a sequence. On the other hand, \cite{dosovitskiy2020image} argued that images can also be formulated as sequences. Flattening an image by cutting it into patches in a sequential order results in a sequence of patches. We refer to this structure as Vision Transformer (ViT), and it has achieved state-of-the-art (SOTA) results in various vision tasks when pre-trained on datasets comprising millions of private images~\cite{sun2017revisiting}. The attention mechanism, lacking the inductive bias for local features present in traditional vision models (such as CNNs), requires more extensive training data to achieve its effects~\cite{vaswani2017attention,d2021convit}. \cite{liu2021swin} proposed the Swin Transformer, a variant of ViT, by adjusting patch sizes to mimic the effects coming from the various receptive fields in CNNs. Similar to how deeper CNN layers increase the receptive field, upper layers in Swin Transformer explore information from wider regions. \cite{mehta2021mobilevit} introduced a groundbreaking ViT structure with reduced learning parameters. While the conventional ViT addressed the absence of inductive bias with a significantly larger model capacity, this proposal implemented it solely through tensor unfold and self-attention. Despite these advancements, exploration for the vision foundation model remained insufficient.

\subsection{ViT-22B}

ViT-22B~\cite{dehghani2023scaling}addresses the absence of a vision foundation model and introduces a new ViT architecture along with parameter settings. Instead of a feed-forward network, it incorporates a parallelized linear network and introduces layer normalization for both key and query to prevent gradient exploding. Special parallelization strategies are required to train the 22B parameters. This research tackles this challenge from two novel perspectives. According to their experimental results, the massive ViT structure demonstrates good performance across most downstream tasks. However, in-depth analysis of this structure and information for practical applications at the local scale are still lacking.
\section{Training ViT-22B on Local}

\begin{table}[]
\centering
\resizebox{0.75\textwidth}{!}{%
\begin{tabular}{cccc|cc|c}
\hline
\multirow{2}{*}{Layers} & \multirow{2}{*}{Hidden D} & \multirow{2}{*}{MLP Size} & \multirow{2}{*}{Heads} & \multicolumn{2}{c|}{\# Params} & \multirow{2}{*}{Note} \\ \cline{5-6}
                        &                           &                           &                        & ViT           & ViT-22B        &                       \\ \hline
12                      & 192                       & 768                       & 3                      & 5.8M          & 5.6M           & Tiny/16               \\
12                      & 384                       & 1536                      & 6                      & 22.2M         & 22.1M          & Small/16              \\
12                      & 768                       & 3072                      & 12                     & 86M           & 87.9M          & Base/16               \\
24                      & 1024                      & 4096                      & 16                     & 307M          & 307M           & Large/16              \\
32                      & 1280                      & 5120                      & 16                     & 632M          & -              & ViT-H/16               \\
48                      & 6144                      & 24576                     & 48                     & -             & 21743M         & ViT-22B/16            \\ \hline
\end{tabular}
}
\caption{Recipe and the number of parameters.}
\label{tab:my-table}
\end{table}

\subsection{Training scheme for ViT-22B}

\begin{align}
    y^{\prime}=\text{LayerNorm}(x)\\
    y=x+\text{LayerNorm}(\text{MLP}(y'))+\text{Attention}(y')\label{eq:2}
\end{align}

\begin{figure}
  \centering
  \begin{subfigure}{\textwidth}
    \centering
    \includegraphics[width=1\linewidth]{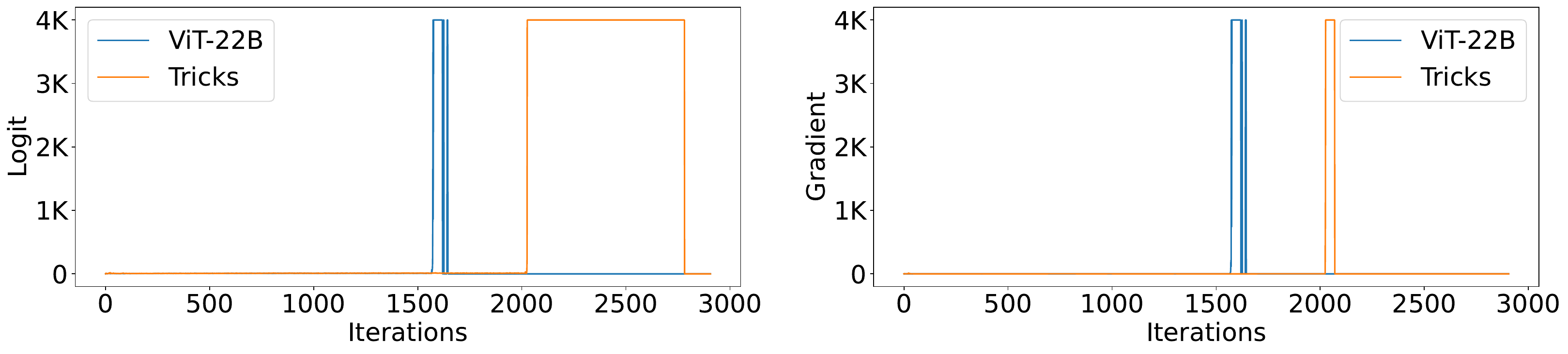}
    \caption{Logits and gradients when original ViT-22B is combined with slight tricks. For convenience, we replaced all values exceeding 4K with 4K for plotting. After the gradient exploding, the optimizer takes very large steps, causing the model to essentially collapse.}
    \label{fig:image_a}
  \end{subfigure}

  \begin{subfigure}{\textwidth}
    \centering
    \includegraphics[width=1\linewidth]{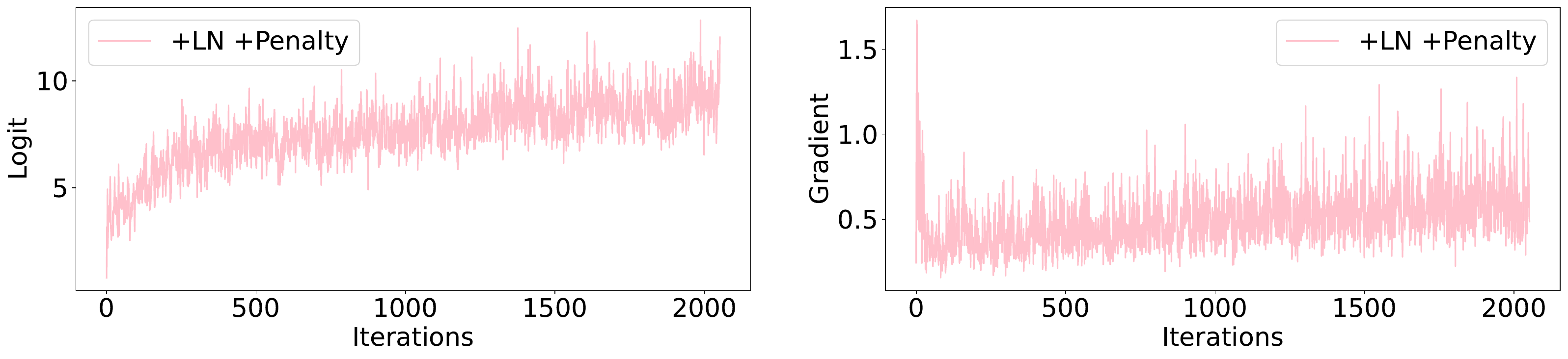}
    \caption{The LayerNorm and gradient penalty to the model enabled stable training for hundreds of epochs.}
    \label{fig:exploding-b}
  \end{subfigure}

  \caption{Logit and gradients between three training strategies.}
  \label{fig:exploding}
\end{figure}

\autoref{fig:exploding} illustrates the unstable training process of ViT-22B with a maximum of 86 million learnable parameters. ViT-22B showed logit values exceeding 4K around 1500 iterations (after 2 epochs). The maximum gradient value also extremely increased, indicating gradient exploding. To mitigate this issue, we employed well-known \textit{Tricks}, including gradient clipping, auto mixed precision~\cite{micikevicius2017mixed}, and a small learning rate. For gradient clipping, we empirically set the threshold to 10 after observing the gradients. Learning rates ranging from $1e-5$ to $5e-4$ were used. However, these conventional tricks only delayed the manifestation of gradient exploding without preventing it. To stabilize the local training of ViT-22B, we employed three methods. Small learning rates and weight decay were used to prevent the explosive growth of gradients. Finally, we made slight modifications to the model's structure. In the case of ViT-22B, linear networks are placed in parallel with self-attention. While in the original ViT-22B, LayerNorm~\cite{ba2016layer} was applied only to query and key for preventing gradient exploding~\cite{dehghani2023scaling, gilmerintriguing}, there was no normalization regulating the explosive gradient flow in parallel linear networks. Therefore, we applied LayerNorm to the output of linear networks as well, as shown in \autoref{eq:2}. After applying these three methods, the training became stable, as depicted in \autoref{fig:exploding-b}, and we could optimize the network up to 200 epochs without encountering gradient exploding.

\subsection{Classification from scratch}

\begin{figure}
  \centering

  \begin{subfigure}{\textwidth}
    \centering
    \includegraphics[width=1\linewidth]{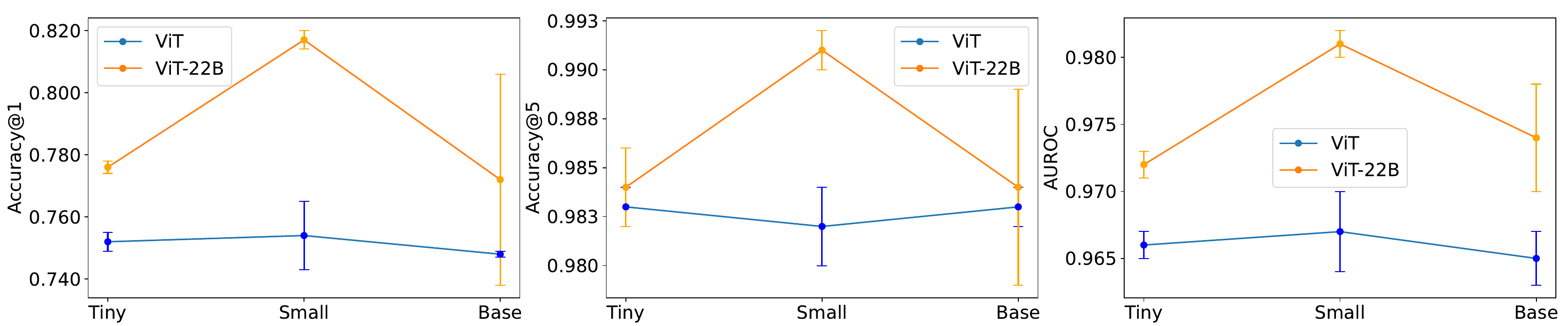}
    \caption{Experiment results on CIFAR-10}
    \label{fig:image_a}
  \end{subfigure}

  \begin{subfigure}{\textwidth}
    \centering
    \includegraphics[width=1\linewidth]{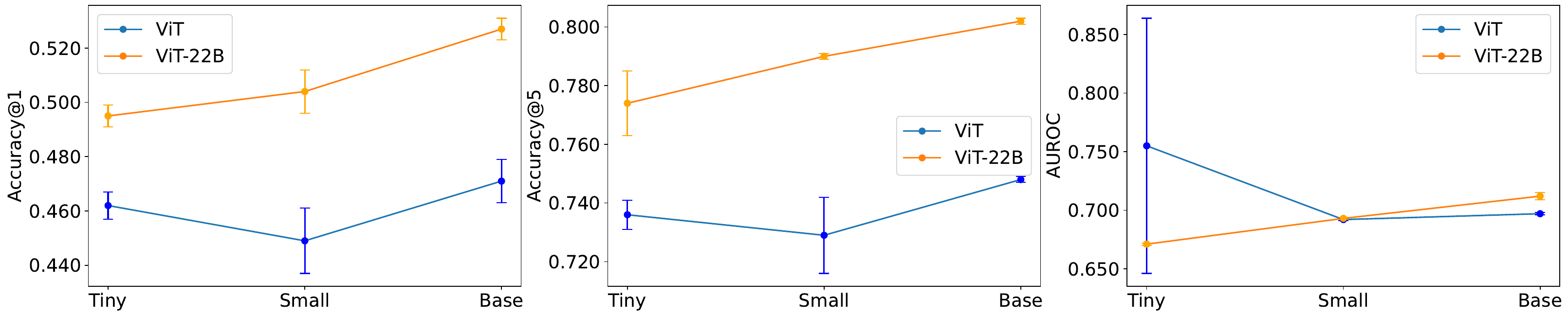}
    \caption{Experiment results on CIFAR-100}
    \label{fig:image_b}
  \end{subfigure}

  \caption{Classification results of ViTs that trained from scratch.}
  \label{fig:cls}
\end{figure}

We conducted an objective comparison of the performance between models trained from scratch, namely ViT-22B and ViT. The experimental results from \cite{dehghani2023scaling} were not objectively comparable due to variations in the sizes of pretraining datasets. Particularly for ViT-22B, being pretrained on a 4B dataset provided a significantly advantageous starting point compared to ViT, which was trained on a 300M dataset. We trained the models using the AdamW optimizer~\cite{loshchilov2017decoupled} for 200 epochs and introduced a multistep learning rate scheduler at 30\%, 60\%, and 90\% of the total epochs. Only elementary spatial augmentations like flips were applied.

\autoref{fig:cls} presents the classification experiment results for CIFAR-10 and 100. Overall, the ViT-22B architecture exhibited superior performance to ViT. In the case of CIFAR-10, models from the Small recipe with fewer parameters outperformed Base recipe models with over 80M parameters. This suggests that when the dataset size is not sufficiently large, a large model may not be necessary for training ViT. On larger datasets like CIFAR-100, there was a tendency for performance to improve with the increase in model size. The results obtained from training from scratch show numerical values similar to the findings\footnote{\url{https://github.com/ra1ph2/Vision-Transformer}}\footnote{\url{https://github.com/kentaroy47/vision-transformers-cifar10}}.

\section{Image Generation}
We explore a topic not investigated in the ViT-22B research: Generative Models. While ViT has been extensively studied and practically applied in almost all areas of vision, it is not commonly used in generative models. This is because ViT's self-attention does not inherently capture the representation of images~\cite{raghu2021vision, battaglia2018relational}. As a result, only a few studies have ventured into image generation based on ViT. \cite{jiang2021transgan} proposed a method for generating high-resolution images from noise z using the ViT structure. This involves upscaling images generated from noise to create high-resolution images. Simultaneously, an auxiliary task for resolution is introduced as a loss term to generate high-quality images. \cite{lee2021vitgan}, similar to \cite{karras2019style}, first maps the style into a certain embedding space and then feeds it into the ViT structure to generate images. \cite{esser2021taming} uses a feature bag called a code book to preserve the inductive bias advantage of CNN. Since these studies either generate images from noise or require a reference, they cannot be applied to image-to-image translation. Therefore, we propose an architecture that enables image-to-image translation using both the ViT and ViT-22B structures.

\subsection{New architecture for ViT based Image2Image translation}

\begin{figure}
    \centering
    \includegraphics[width=1.0\linewidth]{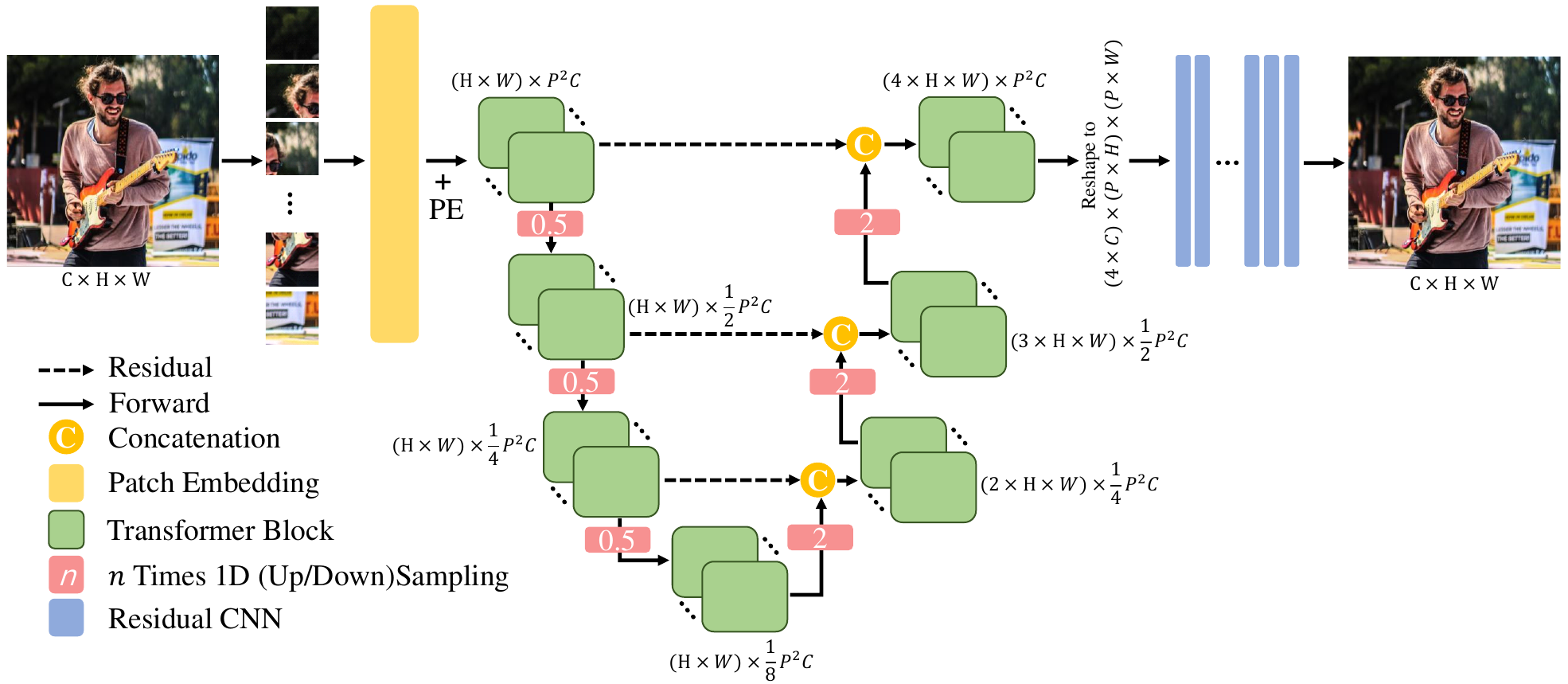}
    \caption{ViTUnet Architecture}
    \label{fig:vitunet}
\end{figure}

% 우리는 이미지 translation을 위한 ViT 기반 생성 구조를 제안한다: ViTUnet. 이 구조는 이미지 생성에서 가장 많이 쓰이고있는 Unet의 구조에서 모티베이션했다. 다양한 resolution을 관찰하는 CNN 기반의 모델의 장점을 흉내낸 X의 구조처럼, 우리의 구조도 다양한 size의 인풋을 받아 이것을 퓨전하는 방식으로 작동한다. 그림 3은 ViTUnet의 전체 플로우를 보여준다. 인풋은 p 사이즈의 패치로 나뉘에 patch dimension으로 임베딩된다. 이 임베딩에 우리는 학습가능한 positional embedding을 더해주어 공간 정보를 제공한다. 모델은 크게 인코딩과 디코딩 과정으로 구성되며 디코딩 과정은 transformer 구조와 CNN residual 구조를 모두 가진다. 인코딩에서 임베딩된 패치는 depth/6개의 transformer 구조를 통과한 후 downsampling 되어 다음 depth/6회 만큼의 transformer 구조에 feed 된다. 이후 디코딩에서는 인풋이 2배만큼 upsampling되어 depth/6 회 만큼의 transformer 구조에 feed된다. 디코딩 과정에서 모든 인풋은 residual 후 concatenation을 포함한다. 이것은 인풋의 패치 개수를 늘리는 방향으로 병합된다. 이후 resahping을 동해서 이미지의 width와 channel을 맞추고 CNN residual을 통해서 원본 이미지의 shape을 재구성한다. 이러한 구조는 ViT가 가진 attention구조를 resizing과 함께 사용할 수 있으며, residual을 통해서 손실되는 정보를 최대한 줄이며 이미지를 재구성할 수 있다.

We propose a ViT-based generative architecture for image translation: ViTUnet. This structure draws inspiration from the widely used Unet~\cite{ronneberger2015u} in image generation. Similar to the Swin~\cite{liu2021swin} architecture, which mimics the advantages of CNN-based models observing various resolutions, our structure operates by receiving inputs of various sizes and fusing them. \autoref{fig:vitunet} illustrates the entire flow of ViTUnet. The input is divided into patches of size $p$, embedded with patch dimension. To this embedding, we add learnable positional embeddings to provide spatial information. The model consists of encoding and decoding processes, with the decoding process incorporating both transformer and CNN residual structures. In the encoding, the embedded patches undergo downsampling after passing through a transformer structure with depth/6 iterations, and they are then fed into the next transformer structure for depth/6 iterations. Subsequently, in the decoding, the input is upsampled by twice as much and fed into a transformer structure for depth/6 iterations. Throughout the decoding process, all inputs include residual concatenation, merging in the direction of increasing the number of input patches. Following reshaping to align the width and channel of the image and reconstructing the shape of the original image through CNN residuals. This structure allows ViT's attention mechanism to be used in conjunction with resizing, minimizing the loss of information through residuals, and enabling the reconstruction of images.

\subsection{Generation results}

\begin{figure}
  \centering
  \begin{subfigure}{\textwidth}
    \centering
    \includegraphics[width=1\linewidth]{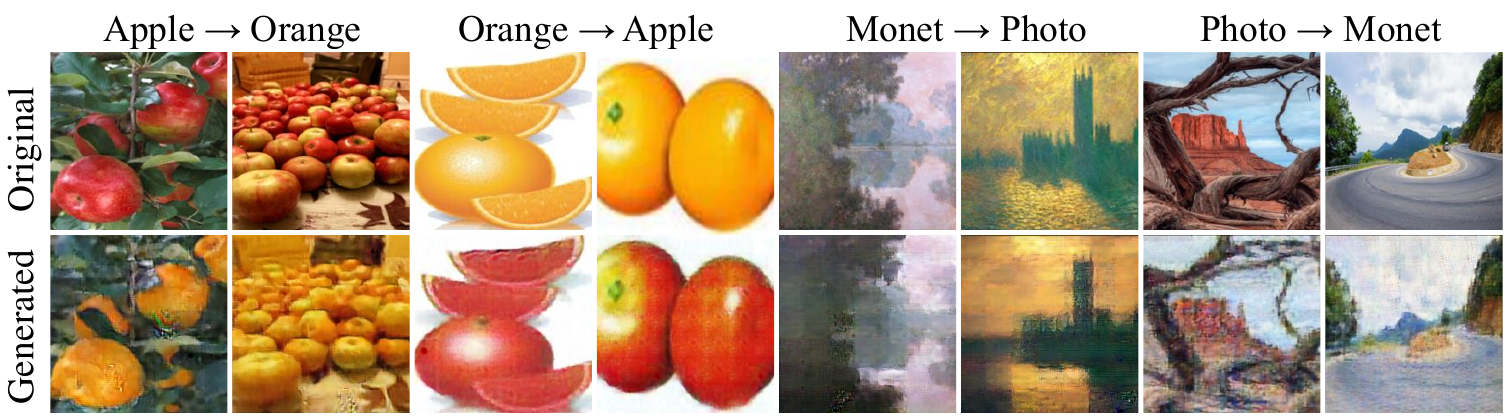}
    \caption{Images that ViTUnet generated.}
    \label{fig:vitunetoutcome_a}
  \end{subfigure}

  \begin{subfigure}{\textwidth}
    \centering
    \includegraphics[width=1\linewidth]{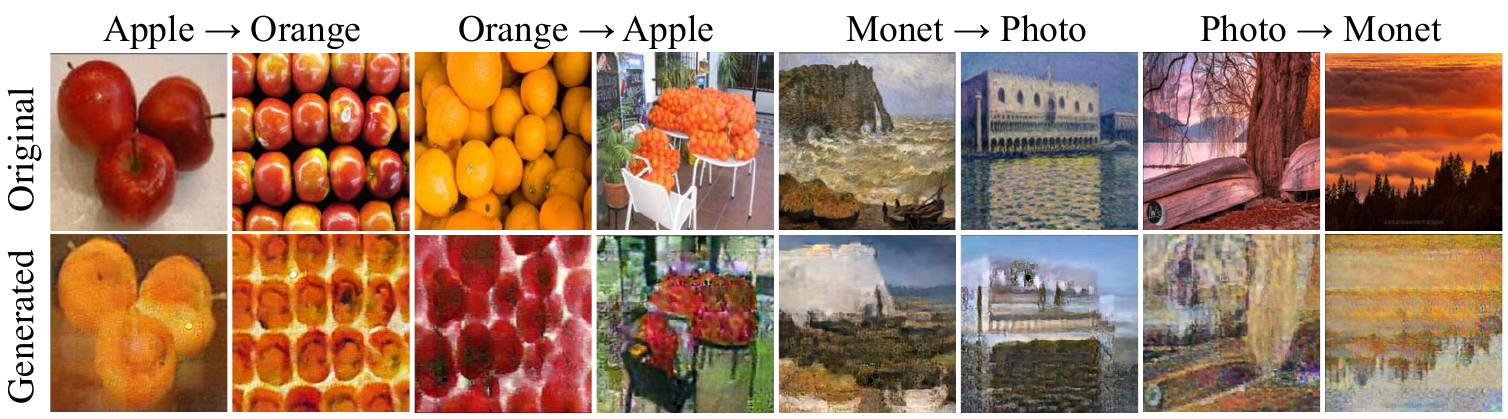}
    \caption{Images that ViT22BUnet generated.}
    \label{fig:vitunetoutcome-b}
  \end{subfigure}

  \caption{Results of image generation for Apple2Orange and Monet2Photo}
  \label{fig:vitunetoutcome}
\end{figure}

\begin{table}[]
\centering
\resizebox{0.75\textwidth}{!}{%
\begin{tabular}{cc|ccc|ccc}
\hline
Backbone    & \# Params & Apple & Orange & Average & Monet & Photo & Average \\ \hline
ViT-S/32    & 12M & \textbf{2.430} & 3.455  & 2.942   & 4.087 & 5.332 & 4.710   \\
ViT-22B-S/32 & 12M & 12.485& 8.503  & 10.494  & 2.357 & 2.638 & 2.498   \\
ViT-B/32    & 44M  & 3.781 & \textbf{1.872}  & \textbf{2.827}   & \textbf{2.154} & \textbf{2.560} & \textbf{2.357}   \\ \hline
\end{tabular}
}
\caption{FID score}
\label{tab:fid}
\end{table}

\begin{figure}
    \centering
    \includegraphics[width=0.8\linewidth]{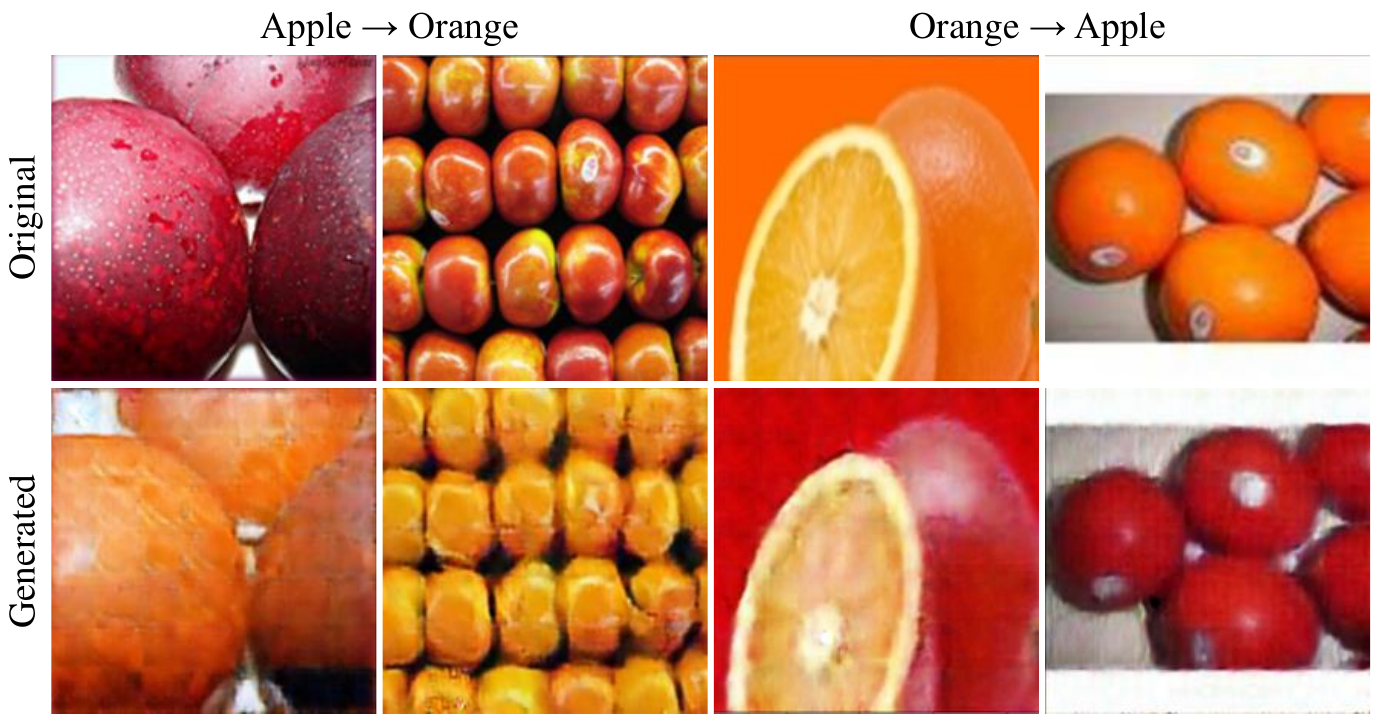}
    \caption{Image generation results from ViT-B/32 backboned ViTUnet. It has 44M parameters.}
    \label{fig:gan_big}
\end{figure}

% 우리는 image-to-image translation으로 Apple2Orange와 Photo2Monet을 사용했으며 X의 training 전략을 그대로 따랐다. D는 70x70 receptiv field를 가지는 CNN patch gan이다. 그림 4는 ViTUnet의 backbone을 ViT 그대로 사용했을 떄와 ViT-22B 구조를 사용했을 때의 생성 결과를 보여준다. 두 backbone 모두 Monet2Photo task에서 정성적으로 나쁜 성능을 보인다. 하지만 사진에서 모네 스타일로의 변경은 비교적 말이 되는 수준의 이미지를 출력했다. 사과와 오렌지의 translation에서는 ViT backbone만이 이해 가능한 수준의 이미지를 출력했다. ViT-22B backbone ViTUnet의 경우 과일 밖의 형상과 색감의 reconstruction에 전체적으로 실패한 모습을 보여준다. 마지막으로 우리는 ViT-B를 backbone으로 사용하여 학습 파라미터를 크게 늘려 이미지를 생성했다. 그림 5는 생성된 이미지를 보여준다. 전체적으로 사과와 오렌지의 스타일 번화를 주었고 그 외의 색감에는 크게 변화가 없다. 다만 패치스러운 형상을 해결하지는 못했다. 이것은 모델의 사이즈가 아닌 ViT의 attention 과정에서 유발되는 것으로 보인다. 테이블2는 이미지 생성 실험의 Fréchet inception distance (FID)를 보여준다. 가장 큰 모델에서 가장 좋은 FID 스코어를 기록했다. 전체적으로 poor한 사과와 오렌지 이미지를 출력했던 ViT-22B backbone 모델이 가장 큰 FID 스코어를 기록했다. 정성적인 평가와 다르게 ViT-22B backbone 모델이 ViT-S backbone 모델보다 Monet2Photo에서 더 좋은 FID 스코어를 받았다. 이것은 생성모델 평가에서 FID 스코어가 꼭 절대적인 기준이 될 수는 없음을 보여준다.

We employed Apple2Orange and Photo2Monet for image-to-image translation, following the training strategy of \cite{zhu2017unpaired}. $D$ is a CNN patch GAN~\cite{isola2017image} with a 70x70 receptive field. \autoref{fig:vitunetoutcome} illustrates the generation results when ViTUnet uses a backbone with ViT-S and when using the ViT-22B-S as a backbone. Both backbones exhibit qualitatively poor performance in the Monet2Photo task. However, the transformation from photo to Monet style yields relatively reasonable images. In the translation of apples and oranges, only the ViT backbone produces images at an understandable level. ViT-22B-S backbone fails overall in the reconstruction of shapes and colors outside the fruits. Lastly, we employed ViT-B as the backbone, increasing the number of training parameters, and generated images. \autoref{fig:gan_big} displays the generated images, providing an overall style transformation for apples and oranges with minimal changes in other color tones. However, it fails to address the patchy shapes, which seem to stem from ViT's attention process rather than the model size. \autoref{tab:fid} presents the Fréchet inception distance (FID) for the image generation experiments, where the largest model achieved the best FID score. Despite generating subpar apple and orange images, the ViT-22B backbone model recorded the highest FID score. In contrast to the qualitative evaluation, the ViT-22B-S backbone model outperformed the ViT-S backbone model in Monet2Photo with a better FID score. This highlights that FID scores may not be an absolute criterion in the evaluation of generative models.
\section{Conclusion}

% 우리는 LLM-like vision model인 ViT-22B를 로컬단에서 학습하여 이 모델에대한 이해를 늘렸다. 로컬에서 이 모델은 불안정한 학습을 보여주었고 우리는 이것을 안정적으로 만들기위한 몇 가지 방법을 제시했다. 우리는 같은 조건에서 학습되었을 경우, ViT-22B가 기존의 ViT 구조의 성능을 상회함을 확인했다.
% 추가적으로 우리는 ViT 구조에서 많이 시도되지 않았던 이미지 생성 모델을 제안했다. 이것은 ViT 구조 역시 이미지 생성에 유용한 도구가 될 수 있음을 보엿다. 하지만 ViT-22B 구조가 이미지 생성에 있어서도 더 좋은 결과를 낸다고 단언할 수는 없었다. 이 결과로 미루어보아 ViT-22B에서 밝히지 않은 실험 결과들에 대한 검증이 이루어진 이후 이 구조가 대부분의 task에서 유용하다고 할 수 있을 것이다. 우리의 시도 후에도 많은 검증이 이루어져서 vision에서 스탠다드를 가지는 모델이 확립되길 기대한다.

In this study, we enhanced our understanding of the locally trained LLM-like vision model, ViT-22B. Training this model locally revealed unstable, and we proposed several methods to stabilize it. Under the same conditions, we confirmed that ViT-22B surpasses the performance of the traditional ViT architecture. Furthermore, we introduced an image generation model, an aspect less explored in ViT structures. This suggests that ViT structures can also serve as valuable tools for image generation. However, we cannot assert that the ViT-22B structure consistently outperforms in image generation. Based on these results, it can be anticipated that, after validation of unexplored experimental outcomes in ViT-22B, this structure might prove beneficial across various tasks. We expect continued validation even after our attempts, paving the way for the establishment of a model setting the standards in vision applications.
% \bibliographystyle{IEEEbib}
% \bibliography{doc/6.Reference.bib}
% \cite{blub}
\bibliographystyle{plain}
\bibliography{main}

\begin{thebibliography}{10}

\bibitem{ba2016layer}
Jimmy~Lei Ba, Jamie~Ryan Kiros, and Geoffrey~E Hinton.
\newblock Layer normalization.
\newblock {\em arXiv preprint arXiv:1607.06450}, 2016.

\bibitem{battaglia2018relational}
Peter~W Battaglia, Jessica~B Hamrick, Victor Bapst, Alvaro Sanchez-Gonzalez, Vinicius Zambaldi, Mateusz Malinowski, Andrea Tacchetti, David Raposo, Adam Santoro, Ryan Faulkner, et~al.
\newblock Relational inductive biases, deep learning, and graph networks.
\newblock {\em arXiv preprint arXiv:1806.01261}, 2018.

\bibitem{bommasani2021opportunities}
Rishi Bommasani, Drew~A Hudson, Ehsan Adeli, Russ Altman, Simran Arora, Sydney von Arx, Michael~S Bernstein, Jeannette Bohg, Antoine Bosselut, Emma Brunskill, et~al.
\newblock On the opportunities and risks of foundation models.
\newblock {\em arXiv preprint arXiv:2108.07258}, 2021.

\bibitem{chen2021generative}
Xiaocong Chen, Yun Li, Lina Yao, Ehsan Adeli, and Yu~Zhang.
\newblock Generative adversarial u-net for domain-free medical image augmentation.
\newblock {\em arXiv preprint arXiv:2101.04793}, 2021.

\bibitem{chowdhery2023palm}
Aakanksha Chowdhery, Sharan Narang, Jacob Devlin, Maarten Bosma, Gaurav Mishra, Adam Roberts, Paul Barham, Hyung~Won Chung, Charles Sutton, Sebastian Gehrmann, et~al.
\newblock Palm: Scaling language modeling with pathways.
\newblock {\em Journal of Machine Learning Research}, 24(240):1--113, 2023.

\bibitem{dehghani2023scaling}
Mostafa Dehghani, Josip Djolonga, Basil Mustafa, Piotr Padlewski, Jonathan Heek, Justin Gilmer, Andreas~Peter Steiner, Mathilde Caron, Robert Geirhos, Ibrahim Alabdulmohsin, et~al.
\newblock Scaling vision transformers to 22 billion parameters.
\newblock In {\em International Conference on Machine Learning}, pages 7480--7512. PMLR, 2023.

\bibitem{dosovitskiy2020image}
Alexey Dosovitskiy, Lucas Beyer, Alexander Kolesnikov, Dirk Weissenborn, Xiaohua Zhai, Thomas Unterthiner, Mostafa Dehghani, Matthias Minderer, Georg Heigold, Sylvain Gelly, et~al.
\newblock An image is worth 16x16 words: Transformers for image recognition at scale.
\newblock {\em arXiv preprint arXiv:2010.11929}, 2020.

\bibitem{d2021convit}
St{\'e}phane d’Ascoli, Hugo Touvron, Matthew~L Leavitt, Ari~S Morcos, Giulio Biroli, and Levent Sagun.
\newblock Convit: Improving vision transformers with soft convolutional inductive biases.
\newblock In {\em International Conference on Machine Learning}, pages 2286--2296. PMLR, 2021.

\bibitem{esser2021taming}
Patrick Esser, Robin Rombach, and Bjorn Ommer.
\newblock Taming transformers for high-resolution image synthesis.
\newblock In {\em Proceedings of the IEEE/CVF conference on computer vision and pattern recognition}, pages 12873--12883, 2021.

\bibitem{fuchs2020se}
Fabian Fuchs, Daniel Worrall, Volker Fischer, and Max Welling.
\newblock Se (3)-transformers: 3d roto-translation equivariant attention networks.
\newblock {\em Advances in neural information processing systems}, 33:1970--1981, 2020.

\bibitem{gilmerintriguing}
Justin Gilmer, Andrea Schioppa, and Jeremy Cohen.
\newblock Intriguing properties of transformer training instabilities, 2023.
\newblock {\em To appear}.

\bibitem{isola2017image}
Phillip Isola, Jun-Yan Zhu, Tinghui Zhou, and Alexei~A Efros.
\newblock Image-to-image translation with conditional adversarial networks.
\newblock In {\em Proceedings of the IEEE conference on computer vision and pattern recognition}, pages 1125--1134, 2017.

\bibitem{jiang2021transgan}
Yifan Jiang, Shiyu Chang, and Zhangyang Wang.
\newblock Transgan: Two transformers can make one strong gan.
\newblock {\em arXiv preprint arXiv:2102.07074}, 1(3), 2021.

\bibitem{karras2019style}
Tero Karras, Samuli Laine, and Timo Aila.
\newblock A style-based generator architecture for generative adversarial networks.
\newblock In {\em Proceedings of the IEEE/CVF conference on computer vision and pattern recognition}, pages 4401--4410, 2019.

\bibitem{lee2021vitgan}
Kwonjoon Lee, Huiwen Chang, Lu~Jiang, Han Zhang, Zhuowen Tu, and Ce~Liu.
\newblock Vitgan: Training gans with vision transformers.
\newblock {\em arXiv preprint arXiv:2107.04589}, 2021.

\bibitem{liu2021blendgan}
Mingcong Liu, Qiang Li, Zekui Qin, Guoxin Zhang, Pengfei Wan, and Wen Zheng.
\newblock Blendgan: Implicitly gan blending for arbitrary stylized face generation.
\newblock {\em Advances in Neural Information Processing Systems}, 34:29710--29722, 2021.

\bibitem{liu2021swin}
Ze~Liu, Yutong Lin, Yue Cao, Han Hu, Yixuan Wei, Zheng Zhang, Stephen Lin, and Baining Guo.
\newblock Swin transformer: Hierarchical vision transformer using shifted windows.
\newblock In {\em Proceedings of the IEEE/CVF international conference on computer vision}, pages 10012--10022, 2021.

\bibitem{loshchilov2017decoupled}
Ilya Loshchilov and Frank Hutter.
\newblock Decoupled weight decay regularization.
\newblock {\em arXiv preprint arXiv:1711.05101}, 2017.

\bibitem{mehta2022large}
Harsh Mehta, Abhradeep Thakurta, Alexey Kurakin, and Ashok Cutkosky.
\newblock Large scale transfer learning for differentially private image classification.
\newblock {\em arXiv preprint arXiv:2205.02973}, 2022.

\bibitem{mehta2021mobilevit}
Sachin Mehta and Mohammad Rastegari.
\newblock Mobilevit: light-weight, general-purpose, and mobile-friendly vision transformer.
\newblock {\em arXiv preprint arXiv:2110.02178}, 2021.

\bibitem{micikevicius2017mixed}
Paulius Micikevicius, Sharan Narang, Jonah Alben, Gregory Diamos, Erich Elsen, David Garcia, Boris Ginsburg, Michael Houston, Oleksii Kuchaiev, Ganesh Venkatesh, et~al.
\newblock Mixed precision training.
\newblock {\em arXiv preprint arXiv:1710.03740}, 2017.

\bibitem{moor2023foundation}
Michael Moor, Oishi Banerjee, Zahra Shakeri~Hossein Abad, Harlan~M Krumholz, Jure Leskovec, Eric~J Topol, and Pranav Rajpurkar.
\newblock Foundation models for generalist medical artificial intelligence.
\newblock {\em Nature}, 616(7956):259--265, 2023.

\bibitem{raghu2021vision}
Maithra Raghu, Thomas Unterthiner, Simon Kornblith, Chiyuan Zhang, and Alexey Dosovitskiy.
\newblock Do vision transformers see like convolutional neural networks?
\newblock {\em Advances in Neural Information Processing Systems}, 34:12116--12128, 2021.

\bibitem{ramesh2021zero}
Aditya Ramesh, Mikhail Pavlov, Gabriel Goh, Scott Gray, Chelsea Voss, Alec Radford, Mark Chen, and Ilya Sutskever.
\newblock Zero-shot text-to-image generation.
\newblock In {\em International Conference on Machine Learning}, pages 8821--8831. PMLR, 2021.

\bibitem{ronneberger2015u}
Olaf Ronneberger, Philipp Fischer, and Thomas Brox.
\newblock U-net: Convolutional networks for biomedical image segmentation.
\newblock In {\em Medical Image Computing and Computer-Assisted Intervention--MICCAI 2015: 18th International Conference, Munich, Germany, October 5-9, 2015, Proceedings, Part III 18}, pages 234--241. Springer, 2015.

\bibitem{sun2017revisiting}
Chen Sun, Abhinav Shrivastava, Saurabh Singh, and Abhinav Gupta.
\newblock Revisiting unreasonable effectiveness of data in deep learning era.
\newblock In {\em Proceedings of the IEEE international conference on computer vision}, pages 843--852, 2017.

\bibitem{thoppilan2022lamda}
Romal Thoppilan, Daniel De~Freitas, Jamie Hall, Noam Shazeer, Apoorv Kulshreshtha, Heng-Tze Cheng, Alicia Jin, Taylor Bos, Leslie Baker, Yu~Du, et~al.
\newblock Lamda: Language models for dialog applications.
\newblock {\em arXiv preprint arXiv:2201.08239}, 2022.

\bibitem{vaswani2017attention}
Ashish Vaswani, Noam Shazeer, Niki Parmar, Jakob Uszkoreit, Llion Jones, Aidan~N Gomez, {\L}ukasz Kaiser, and Illia Polosukhin.
\newblock Attention is all you need.
\newblock {\em Advances in neural information processing systems}, 30, 2017.

\bibitem{wang2023internimage}
Wenhai Wang, Jifeng Dai, Zhe Chen, Zhenhang Huang, Zhiqi Li, Xizhou Zhu, Xiaowei Hu, Tong Lu, Lewei Lu, Hongsheng Li, et~al.
\newblock Internimage: Exploring large-scale vision foundation models with deformable convolutions.
\newblock In {\em Proceedings of the IEEE/CVF Conference on Computer Vision and Pattern Recognition}, pages 14408--14419, 2023.

\bibitem{yi2019generative}
Xin Yi, Ekta Walia, and Paul Babyn.
\newblock Generative adversarial network in medical imaging: A review.
\newblock {\em Medical image analysis}, 58:101552, 2019.

\bibitem{zhai2022scaling}
Xiaohua Zhai, Alexander Kolesnikov, Neil Houlsby, and Lucas Beyer.
\newblock Scaling vision transformers.
\newblock In {\em Proceedings of the IEEE/CVF Conference on Computer Vision and Pattern Recognition}, pages 12104--12113, 2022.

\bibitem{zhu2017unpaired}
Jun-Yan Zhu, Taesung Park, Phillip Isola, and Alexei~A Efros.
\newblock Unpaired image-to-image translation using cycle-consistent adversarial networks.
\newblock In {\em Proceedings of the IEEE international conference on computer vision}, pages 2223--2232, 2017.

\end{thebibliography}
% \bibliography{mybibfile}
% \bibliography{doc/6.Reference.bib}

\end{document}